\pdfoutput=1

\documentclass[11pt]{article}

\usepackage{acl}

\usepackage{times}
\usepackage{latexsym}
\usepackage{graphicx}
\usepackage{subfigure}
\usepackage{multirow}

\usepackage[T1]{fontenc}

\usepackage[utf8]{inputenc}

\usepackage{microtype}

\title{Key Information Extraction in Purchase Documents \\ using Deep Learning and Rule-based Corrections}

\author{Roberto Arroyo, Javier Yebes, Elena Mart\'inez, H\'ector Corrales, Javier Lorenzo \\
  NielsenIQ Spain \\
  \footnotesize\texttt{\{roberto.arroyo,javier.yebes,elena.martinez,hector.corrales,javier.lorenzo\}@nielseniq.com} \\}

\begin{document}
\maketitle
\begin{abstract}

Deep Learning (DL) is dominating the fields of Natural Language Processing (NLP) and Computer Vision (CV) in the recent times. However, DL commonly relies on the availability of large data annotations, so other alternative or complementary pattern-based techniques can help to improve results. In this paper, we build upon Key Information Extraction (KIE) in purchase documents using both DL and rule-based corrections. Our system initially trusts on Optical Character Recognition (OCR) and text understanding based on entity tagging to identify purchase facts of interest (e.g., product codes, descriptions, quantities, or prices). These facts are then linked to a same product group, which is recognized by means of line detection and some grouping heuristics. Once these DL approaches are processed, we contribute several mechanisms consisting of rule-based corrections for improving the baseline DL predictions. We prove the enhancements provided by these rule-based corrections over the baseline DL results in the presented experiments for purchase documents from public and NielsenIQ datasets.

\end{abstract}

\section{Introduction}

The intersection between NLP and CV algorithms is a key factor in systems that require processing visual and textual features. There are several use cases in the retailing industry in which this combination is typically applied, such as automated item coding~\citep{ref:Arroyo19wcvpr}, classification of promotions in digital leaflets~\citep{ref:Arroyo20wcoling} or the application of Visual Question Answering (VQA) to store observation systems~\citep{ref:Arroyo22mtap}, among others. In general terms, a great part of these use cases rely on KIE for automatically obtaining data of interest from varied sources related to images and documents. One of the main approaches based on KIE for retail and consumer measurement is focused on the automated recognition of data from purchase documents.

\begin{figure} [!ht] 
\includegraphics[width=\columnwidth]{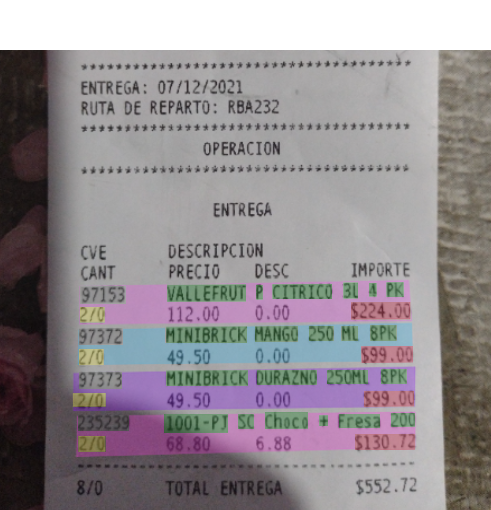}
\caption{Example of KIE in purchase documents from the NielsenIQ dataset. We can see entities tagged as product codes in gray, descriptions in green, quantities in yellow and total prices in red. Moreover, the regions corresponding to a same product jointly with their linked entities are marked in a blue to purple scale.} 
\label{fig:kie_in_purchase_documents} 
\end{figure}

The information of interest commonly acquired by purchase decoding systems is associated with the purchased products printed in the documents to be recognized at item level, as depicted in Fig.~\ref{fig:kie_in_purchase_documents}. Here, there are several entities of text related to the characteristics of a purchased product that are generally extracted, such as the following ones:

\begin{itemize}
 \item \emph{Description}: textual specification for representing a purchased product. 
 \item \emph{Code}: unique numerical identifier of a purchased product.
 \item \emph{Quantity}: total purchased units of the same product.
 \item \emph{Price}: total value of all the purchased units from the same product.
\end{itemize}

In this paper, we propose a novel approach for KIE in purchase documents using DL and \mbox{rule-based} corrections. The presented DL architecture initially extracts the words contained in the document using OCR facilities. The recognized words jointly with the processed image are used as input of an entity tagging model based on Transformers and Convolutional Neural Networks (CNNs), which is able to predict the facts associated with the purchased products. Then, we compute a line detection method based on GNNs (Graph Neural Networks) with the aim of grouping all the text lines corresponding to a same product for entity linking. Finally, our pipeline applies some \mbox{rule-based} refinements to the previously predicted tags to correct possible inconsistencies. 

We observed that standard DL proposals for entity tagging decrease their performance in complex purchase documents with low image quality, so false positives and false negatives commonly appear within the set of predictions. However, we know some specific business rules related to purchase documents for product descriptions, codes, quantities and prices that can be applied in post-processing, with the aim of enhancing the global accuracy of the system. In this way, we can take advantage of the patterns learned by DL and combine them with the human knowledge about the rules related to the purchase document decoding use case, so both can be applied in a complementary way to improve the quality of our whole solution.

In this regard, the main contributions derived from the research presented in this paper are the following ones:

\begin{itemize}
 \item Design of a DL approach for KIE focused on purchase documents, which fuses NLP and CV in an architecture composed of OCR, entity tagging and product line grouping for entity linking.
 \item Definition of rule-based corrections in order to enhance the initial predictions given by the DL approach, demonstrating how \mbox{pattern-based} techniques can complement DL to improve performance.
 \item Presentation of a set of experiments in public and NielsenIQ datasets with the aim of validating our DL approach for KIE in purchase documents, jointly with the enhancements provided by rule-based corrections.
\end{itemize}

The contents of the paper are organized as follows: a review of the main \mbox{state-of-the-art} methods for KIE in document decoding is presented in Section~\ref{sec:related_work}. Our approach for KIE in purchase documents using DL and rule-based corrections is detailed in Section~\ref{sec:our_approach}. The main experiments and results related to our DL proposal and the accuracy improvements of \mbox{rule-based} corrections are discussed in Section~\ref{sec:experiments}. The final conclusions associated with our research and some insights about future works are finally summarized in Section~\ref{sec:conclusions}.

\begin{figure*}[!ht]
\centering
\subfigure[Input.]                                     
{\includegraphics[width=0.19\textwidth]{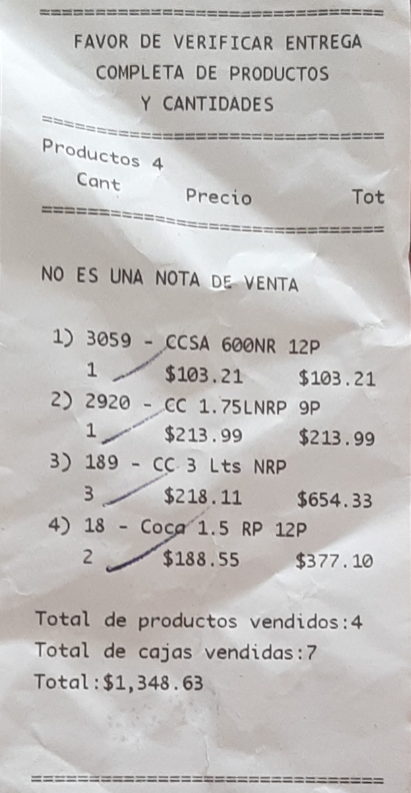}}
\subfigure[OCR.]                                     
{\includegraphics[width=0.19\textwidth]{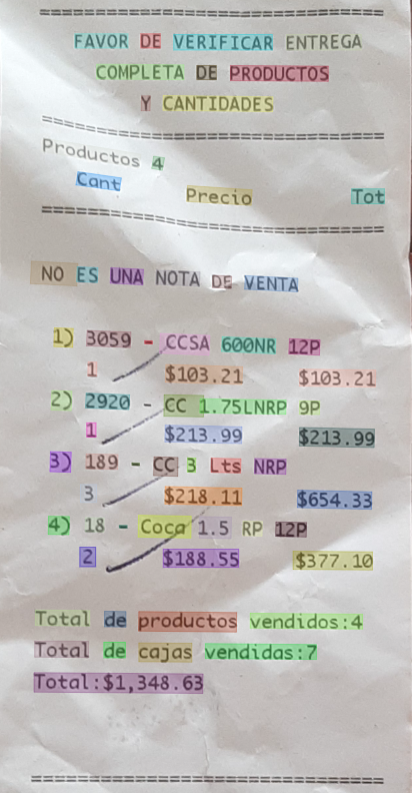}}
\subfigure[Entity tagging.]                                     
{\includegraphics[width=0.19\textwidth]{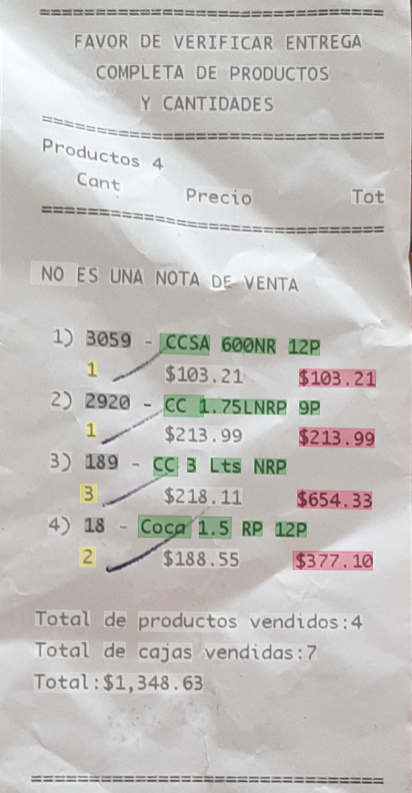}}
\subfigure[Prod. line grouping.]                                     
{\includegraphics[width=0.19\textwidth]{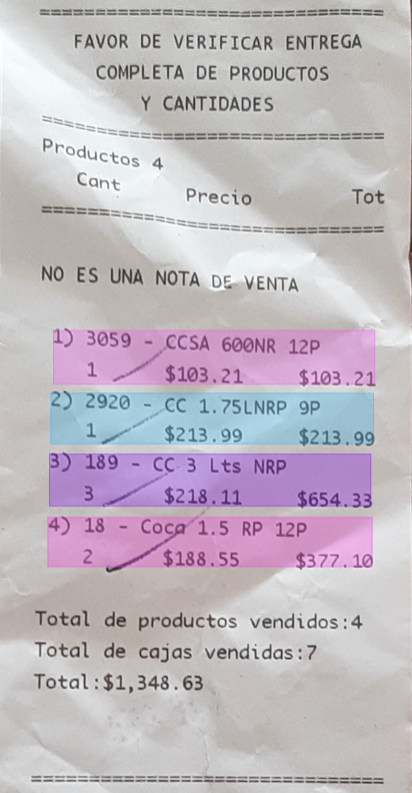}}
\subfigure[Output.]                                     
{\includegraphics[width=0.19\textwidth]{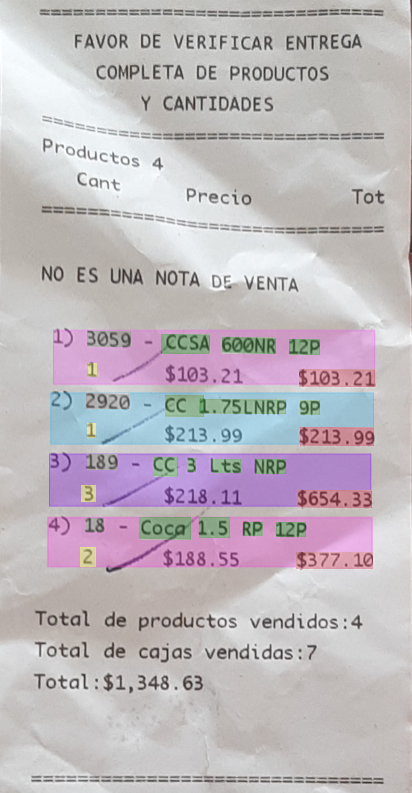}}
\caption{Main stages of our approach for KIE in purchase documents using DL and rule-based corrections. a)~Input:~initial data acquisition and image digitalization. b) OCR: detection and recognition of the words in the image. c) Entity tagging: identification of purchase facts of interest (e.g., product codes, descriptions, quantities, or prices). d) Product line grouping: identification of text lines and grouping by products of interest. e) Output: final predictions after rule-based corrections (if any). In this case, the diagram shows an example were DL was enough to properly predict all the tags, but we will depict specific cases along the paper in which rule-based corrections are carried out with respect to the original entity tagging predictions.}
\label{fig:stages} 
\end{figure*}

\section{Related Work} 
\label{sec:related_work}

The rise of DL has provided powerful tools to the fields of NLP and CV, which are the base of KIE systems focused on purchase documents decoding. This research line has also grown thanks to the proliferation of related public datasets, such as CORD~\citep{ref:Park19wneurips}, SROIE~\citep{ref:Huang19icdar} or FUNSD~\citep{ref:Jaume19wicdar}. Moreover, companies working on retail intelligence also generate large amounts of data that are typically used for researching in real use cases, such as NielsenIQ.

In the intersection between NLP and CV, OCR techniques commonly represent the starting point for recognizing the text contained in a document or image. Methods based on OCR examine images pixel by pixel, looking for shapes that match the character traits. The state of the art in OCR comprises solutions that are open source and proprietary. Tesseract OCR is one of the most effective open-source approaches~\citep{ref:Anwar22springer}. However, proprietary solutions such as Amazon or Google OCR are currently obtaining much better results in text recognition~\citep{ref:Hegghammer22springer}. 

Once the text is recognized, purchase document decoding systems typically need to understand certain parts or extract specific information. In this regard, the recent popularization of Transformers~\citep{ref:Vaswani17neurips} and architectures such as BERT~\citep{ref:Devlin18naacl} in the NLP community has provided new tools for achieving the desired results.

Among the different possibilities of text understanding, document decoding systems for purchase facts are commonly associated with entity tagging. This approach is applied with the aim of extracting specific product information in purchase documents, such as product description, code, quantity or price. Entity tagging is another state-of-the-art field that has been benefited by Transformers. For instance, the architecture defined by PICK~\citep{ref:Yu21icpr} is typically used in entity tagging systems, because its combination of Transformers to get text embeddings and CNNs to obtain image embeddings~\citep{ref:He16cvpr} provides one of the top performances in the recent literature. Another popular technique for entity tagging is LayoutLM and its different versions~\citep{ref:Xu20kdd,ref:Xu21ijcnlp,ref:Huang22arxiv}, including an approach focused on multilingual capabilities~\citep{ref:Xu21arxiv}.

Apart from entity tagging, entity linking is also typically required to match each purchase tag with each respective product in the purchase document. Here, state-of-the-art techniques such as SPADE~\citep{ref:Hwan21ijcnlp} or BROS~\citep{ref:Hong22aaai} propose end-to-end entity linking approaches. However, our proposal considers a method based on two stages, consisting of an initial entity tagging which is then linked by using line detection algorithms and rule-based heuristics for product lines grouping. For line detection, we take advantage of recent GNN proposals such as~\citep{ref:Qasim19icdar}~or~\citep{ref:Carbonell21icpr}. 

Unfortunately, the state of the art does not consider a lot of solutions for correcting wrong entity tagging predictions, which is something typical in complex documents. For this reason, we propose novel rule-based corrections for improving automated entity tagging in purchase documents.

\section{Our Approach}
\label{sec:our_approach}

In this section, our whole pipeline proposed for KIE in purchase documents using DL and \mbox{rule-based} corrections is described. The core architecture builds upon DL in order to obtain the initial predictions for recognizing the text in the documents and tag words into varied categories from the different purchased products characteristics. Once these baseline predictions are provided by the DL pipeline, the implemented rule-based corrections are applied to refine the final output. We explain both DL and rule-based schemas with the aim of fully understanding the complementary solutions designed to extract the information of interest from purchase documents.

\subsection{DL Architecture}
\label{sec:dl_architecture}

The pipeline based on DL is composed of three main stages presented in Fig.~\ref{fig:stages} from input to output. These stages are basically OCR, entity tagging and product line grouping. 

We explain the most common technical challenges for each stage along this section, jointly with their implementations mainly based on DL approaches. The goal is to understand how the learned textual and visual features are impacting the performance of the final system before checking how rule-based patterns can complement them to enhance the final output.

\subsubsection{OCR}
\label{sec:ocr}

The proper selection of a robust OCR engine is crucial for the development of any document decoding system, because it is the bottleneck for the subsequent stages in the DL architecture.

OCR converts purchase images into machine-readable text data. The human visual system reads text by recognizing the patterns of light and dark, translating those patterns into characters and words, and then attaching meaning to it. Similarly, OCR attempts to mimic our visual system by using DL.

As discussed in Section~\ref{sec:related_work}, papers focused on benchmarking OCR engines such as~\citep{ref:Hegghammer22springer} have demonstrated that propietary solutions are currently yielding much accurate results than open-source proposals. Then, we decided to build upon the OCR provided by the Google Vision API\footnote{\url{https://cloud.google.com/vision}}, which provides some of the most remarkable results in text recognition. It must be noted that the goals of this paper are not focused on contributing a new OCR engine, so we decided just to apply Google Vision API features for recognizing text as a base tool for the rest of the stages implemented for KIE in purchase documents using DL.

\begin{figure}[!ht] 
\includegraphics[width=\columnwidth]{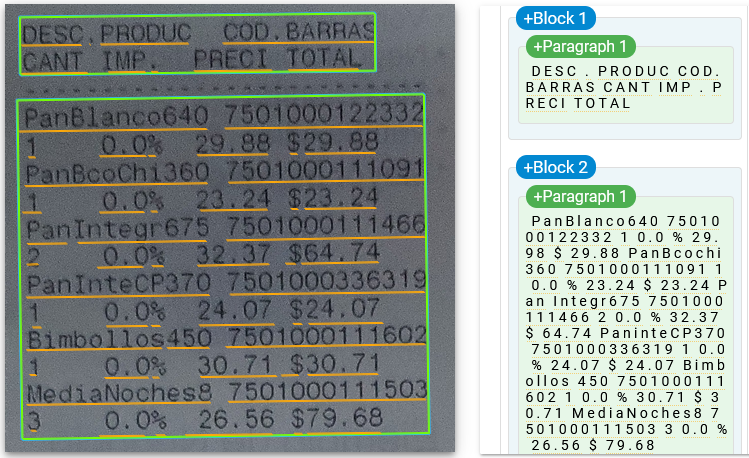}
\caption{Visual example of OCR processing using the Google Vision API over a purchase document. Image is focused on a section where purchased products are printed. The output shows the detected characters, words and paragraphs over the image and the recognized text.}
\label{fig:ocr} 
\end{figure}

The output obtained by the OCR service is composed of the recognized text over the images and the locations of the detected characters, words and paragraphs, as shown in the visual example presented in Fig.~\ref{fig:ocr}. This output is used as input of the subsequent stage for entity tagging in order to classify the different words into their corresponding categories related to the varied purchased products characteristics.

\subsubsection{Entity Tagging}
\label{sec:entity_tagging}

KIE systems typically require algorithms to understand the recognized text and categorize some parts of it. In this sense, entity tagging is used with the aim of categorizing information of interest from purchased products, such as descriptions, codes, quantities or prices.

\begin{figure}[!ht] 
\includegraphics[width=\columnwidth]{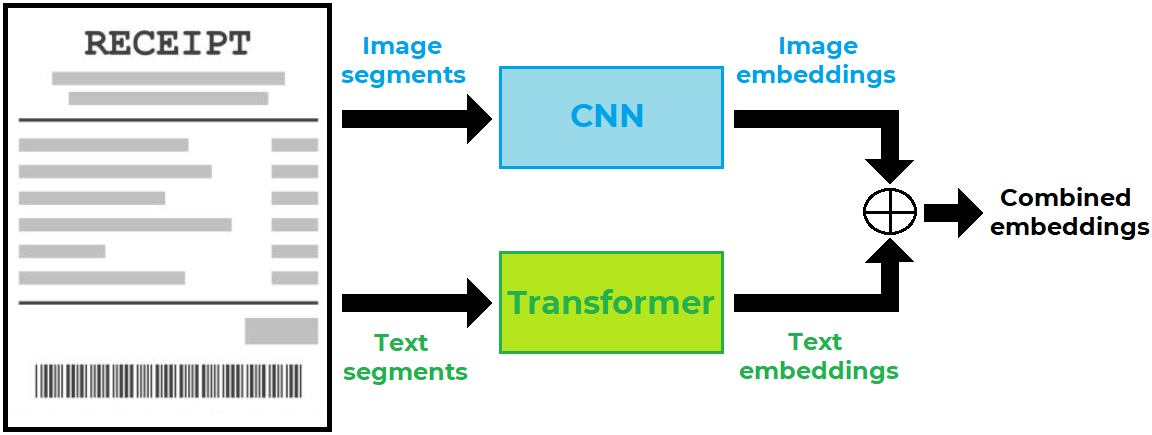}
\caption{Base architecture for entity tagging using DL. Tranformers and CNNs are applied for extracting text and image embeddings, respectively.}
\label{fig:entity-tagging} 
\end{figure}

There are different proposals in the recent state of the art for computing entity tagging based on DL. In our pipeline, we follow an encoder-decoder architecture similar to the presented in works such as~\citep{ref:Yu21icpr}. Within this schema, text features are acquired by a Transformer and image features are processed by a CNN, as shown in Fig.~\ref{fig:entity-tagging}.

The combined embeddings ($CE$) are obtained by fusing the image ($IE$) and text ($TE$) embeddings using an element-wise addition operation, as formulated in Eq.~\ref{eq:entity-tagging_1}. Previously, $IE$ and $TE$ are outputted by the encoders into a vector derived from the CNN~($ie^{(x)}$) and Transformer~($te^{(x)}$) features respectively, as shown in Eq.~\ref{eq:entity-tagging_2} and Eq.~\ref{eq:entity-tagging_3}. The encoders ($\theta$) in each case are defined in Eq.~\ref{eq:entity-tagging_4} and Eq.~\ref{eq:entity-tagging_5}, where $is^{(x)}$ represents the input image segments and $ts^{(x)}$ the corresponding text segments.

\begin{equation}
\label{eq:entity-tagging_1}
CE = IE \oplus TE
\end{equation}
\begin{equation}
\label{eq:entity-tagging_2}
IE = [ie^1, ie^2, ..., ie^{N-1}, ie^N]
\end{equation}
\begin{equation}
\label{eq:entity-tagging_3}
TE = [te^1, te^2, ..., te^{N-1}, te^N]
\end{equation}
\begin{equation}
\label{eq:entity-tagging_4}
ie^{(x)} = \mbox{cnn} (is^{(x)}, \theta_{cnn})
\end{equation}
\begin{equation}
\label{eq:entity-tagging_5}
te^{(x)} = \mbox{transformer} (ts^{(x)}, \theta_{transformer})
\end{equation}

It must be noted that we decided to use a DL encoder-decoder architecture based on Transformers and CNNs because it is one of the most successful approaches in the state of the art, as we will discuss in the experiments presented in Section~\ref{sec:experiments}. However, our subsequent rule-based corrections over DL predictions could be adapted to any other entity tagging technique focused on similar DL schemas.

\subsubsection{Product Line Grouping}
\label{sec:product_line_grouping}

Once entity tags are initially recognized for a purchase document, they are grouped by product to know their relationships and association with specific purchase characteristics. This entity linking is performed by applying a product line grouping step that uses DL and some grouping heuristics.

Firstly, the lines of the document must be individually detected. To do this, a GNN based on \citep{ref:Qasim19icdar} is implemented with the aim of connecting the different words previously extracted as entity tags of interest in a same line. This GNN proposes an architecture based on graph networks that combines the benefits of CNNs for visual feature extraction and graph networks for dealing with the problem structure.

\begin{figure}[!ht]
\centering
\subfigure[Line detection.]                                     
{\includegraphics[width=0.49\columnwidth]{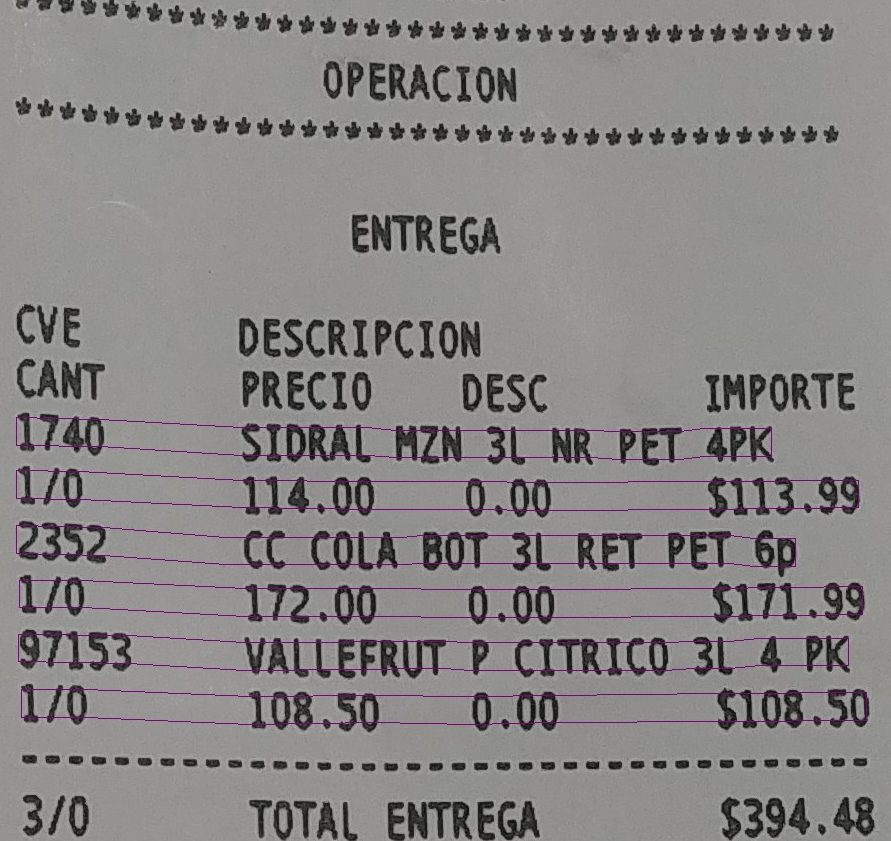}}
\subfigure[Product line grouping.]                                     
{\includegraphics[width=0.49\columnwidth]{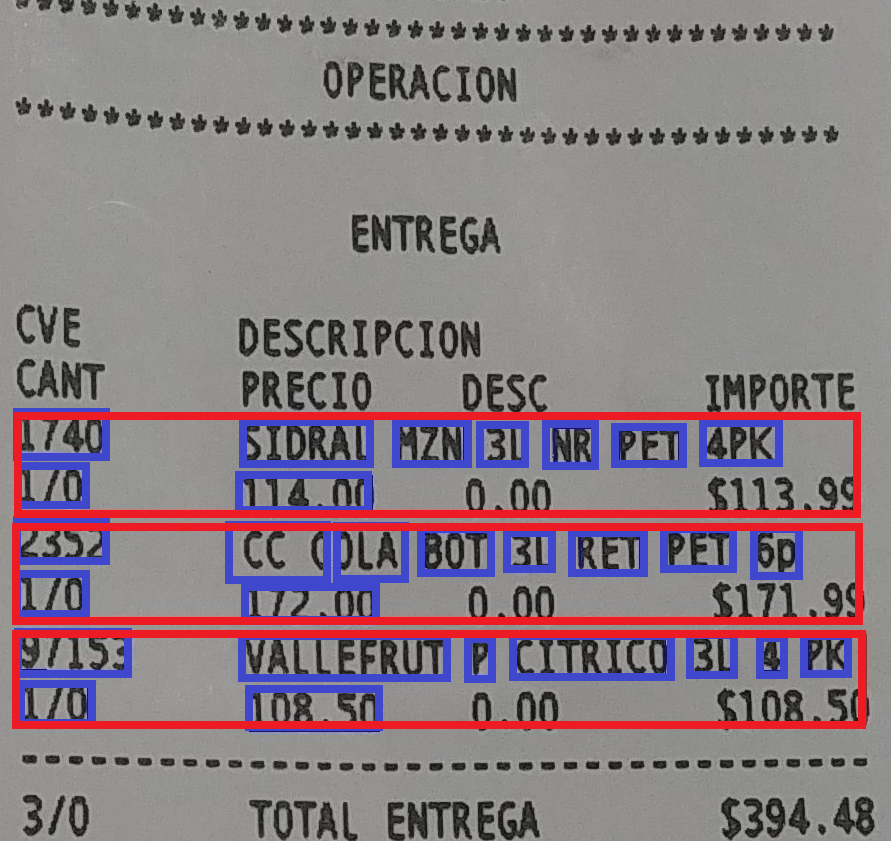}}
\caption{Visual example from line detection to product line grouping. On the left image, the individual lines detected by the GNN from entities of interest are represented in purple. On the right image, the final product line groups are depicted in red, jointly with the entity tagging words of interest in blue.}
\label{fig:lines} 
\end{figure}

After detecting the individual lines, some heuristics are applied to group the lines corresponding to a same product, as illustrated in Fig.~\ref{fig:lines}. The steps of this grouping process are the following ones:

\begin{enumerate}
 \item Take the first individual line detected (from top to bottom in Y axis) and check if there are product description entities. If not, continue searching until the grouping algorithm finds the first line with a product description.
 \item Once the algorithm finds the line with product description entities, there are two options:
  \begin{enumerate}
  \item If the line also contains a product quantity and price (codes are not always available according to business rules), all the information of interest is located in that line and can be individually grouped.
  \item If the line only contains description tags, go to step 3.
  \end{enumerate}
 \item Check if the next line contains another description or any other tag, there are two options:
  \begin{enumerate}
  \item If there are only description tags, it means that the whole product description covers more than one line. The algorithm will search until it finds a line containing any other entity different from description before closing the group of lines.
  \item If there are tags such as product quantity or price, the group of lines can be directly closed. Then, the algorithm takes the previously accumulated lines for the current product and combines their bounding boxes to obtain the final product line grouping.
  \end{enumerate}
\end{enumerate}

\subsection{Rule-based Corrections}
\label{sec:rule-based_corrections}

Current solutions for document decoding trust on DL for varied algorithms related to NLP and CV, as we previously reviewed in our architecture for KIE in purchase documents. Unfortunately, the high variety of formats, qualities and languages in these documents makes difficult to train models able to generalize to every situation. Then, we can expect some errors in the baseline entity tagging predictions provided by a DL pipeline. These errors are not always possible to correct, because in some cases the associated text is very difficult to recognize and understand. However, there are some other cases in which we can apply business rules and human knowledge to find and ameliorate DL errors in predictions.

It is possible to correct some of the missed elements in the initial tags (false negatives) and even the ones that are wrongly detected (false positives) to replace them with the correct ones. Product descriptions are the core of our product line grouping heuristics because they are generally the most accurate in automation and usually contain several words. In the proposed rule-based corrections for entities, we focus on tags that are commonly composed of only one word based on numbers, which are a typical source of issues in purchase document decoding. These entity tags are mainly product codes, product quantities and product prices. It must be noted that although we focus on these common purchase tags, the rule-based corrections presented in this paper could be adapted or updated for other purchase tags depending on the requirements of a specific use case.

\subsubsection{Corrections in Product Codes}
\label{sec:corrections_in_product_codes}

Product codes do not always appear as part of the product information in purchase documents. However, when they are included among these data, they are typically the highest integer number in a product line grouping, as can be seen in the example presented in Fig.~\ref{fig:rules_example_code}. Then, in case we cannot find the product code with the initial DL-based entity tagging, we can re-check if it is available by applying the rule formulated in Eq.~\ref{eq:word_vector} and Eq.~\ref{eq:rule_codes}, where $TW_{pr}$ is a vector composed of all the words ($tw^{(x)}_{pr}$) that are not tagged inside a specific product group. It must be noted that $code_{pr}$ will be only corrected if the number resulting from applying Eq.~\ref{eq:rule_codes} is higher than the minimum integer value among all the remaining words that are not tagged from DL predictions.

\begin{equation}
\label{eq:word_vector}
TW_{pr} = [tw^{(1)}_{pr}, tw^{(2)}_{pr}, ..., tw^{(N-1)}_{pr}, tw^{(N)}_{pr}]
\end{equation}
\begin{equation}
\label{eq:rule_codes}
code_{pr} = \mbox{max}(\mbox{integer}(TW_{pr}))
\end{equation}
\vspace{1px}

In this case, we just try to find the missing code for a product if it has not been previously detected in order to solve false negatives, but it could be also applied for correcting false positives from DL predictions if required. In that scenario, low confidences associated with DL predictions in entity tagging could help to find candidates to be corrected. However, we decided to fix only false negatives and trust on DL for the rest of predictions, with the aim of obtaining the highest benefit from both DL predictions and rule-based corrections, without an implicit dependency on confidence thresholds. In fact, this schema obtains a remarkable enhancement in performance according to the results presented in Section~\ref{sec:results}, without adding the extra complexity and possible issues related to also modifying explicit DL predictions. A similar way of correcting false negatives is proposed for product quantities and prices.

\subsubsection{Corrections in Product Quantities}
\label{sec:corrections_in_product_quantities}

Product quantity is another entity tag typically printed in purchase documents that can be fixed in several cases during post-processing by applying rule-based corrections. In this regard, we must consider that this value usually represents an integer number close or equal to one. 

\begin{figure*}[!ht]
\centering
\subfigure[Baseline entities.]                                     
{\includegraphics[width=0.28\textwidth]{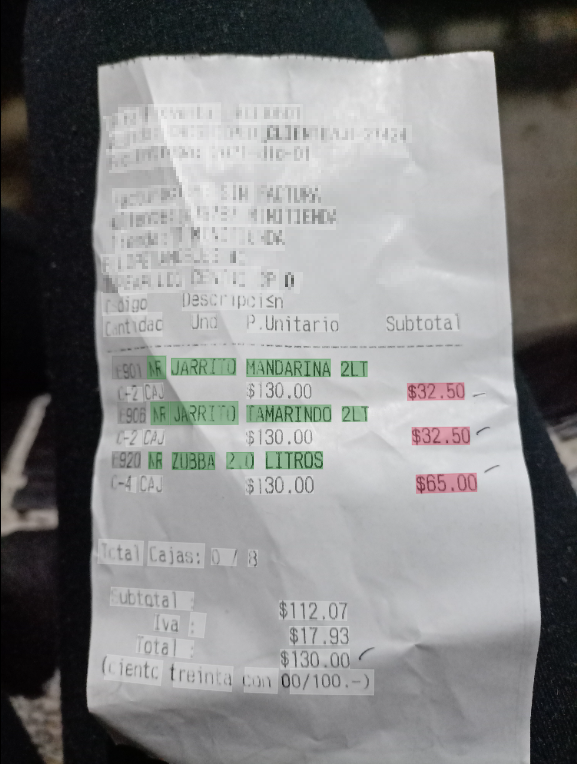}}
\subfigure[Product line grouping.]                                     
{\includegraphics[width=0.28\textwidth]{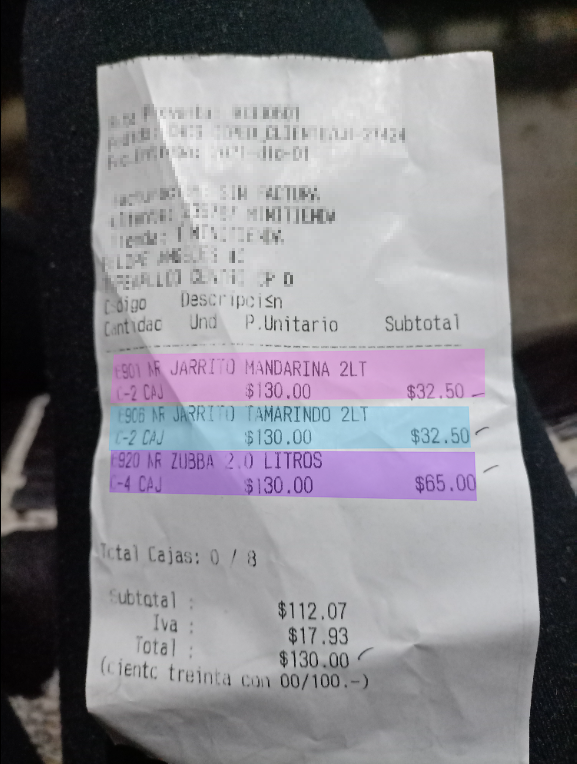}}
\subfigure[Corrected entities for quantities.]                                     
{\includegraphics[width=0.28\textwidth]{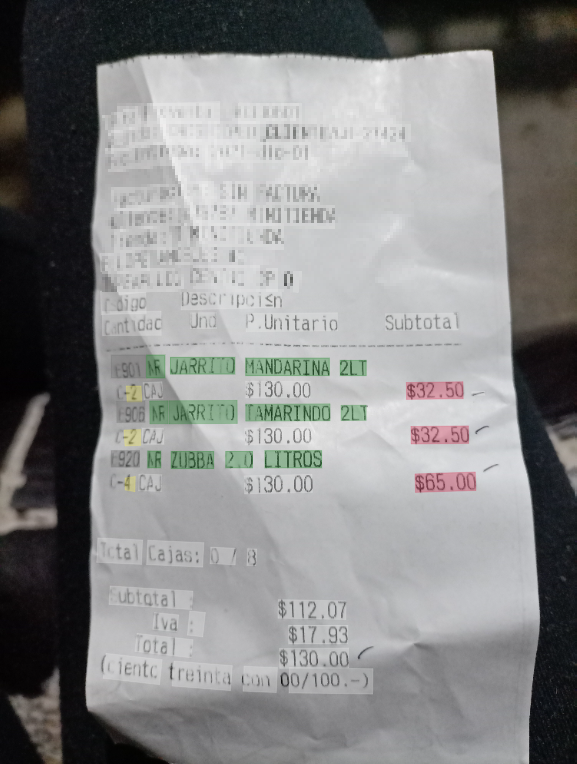}}
\caption{Visual examples of product quantity correction (in yellow) based on rule-based heuristics.}
\label{fig:rules_example_quantity} 
\end{figure*}

According to the previous considerations, we can search for the lowest integer number in a product line grouping in case we did not initially find this tag with DL-based entity tagging, as depicted in the visual examples that are presented in Fig.~\ref{fig:rules_example_quantity}.

More formally, the definition of the rule associated with this correction for missed product quantities is formulated in Eq.~\ref{eq:rule_quantities}. It must be noted that $quantity_{pr}$ will be only corrected if the number resulting from applying Eq.~\ref{eq:rule_quantities} is lower than the maximum integer value among all the remaining words that are not tagged from DL predictions.

\vspace{-5px}
\begin{equation}
\label{eq:rule_quantities}
quantity_{pr} = \mbox{min}(\mbox{integer}(TW_{pr}))
\end{equation}
\vspace{1px}

\subsubsection{Corrections in Product Prices}
\label{sec:corrections_in_product_prices}

Product prices are commonly represented by float numbers. In the use case exposed along this paper, we specifically tag the total prices per product, so it is expected to find the highest float number in these situations. Then, if DL predictions are not able to find a total price for a product line grouping, the rule-based correction formulated in Eq.~\ref{eq:rule_price} is applied to check if it is possible to find a word that is not tagged fulfilling these requirements.

\begin{equation}
\label{eq:rule_price}
price_{pr} = \mbox{max}(\mbox{float}(TW_{pr}))
\end{equation}
\vspace{1px}

\begin{figure*}[!ht]
\centering
\subfigure[Baseline entities.]                                     
{\includegraphics[width=0.28\textwidth]{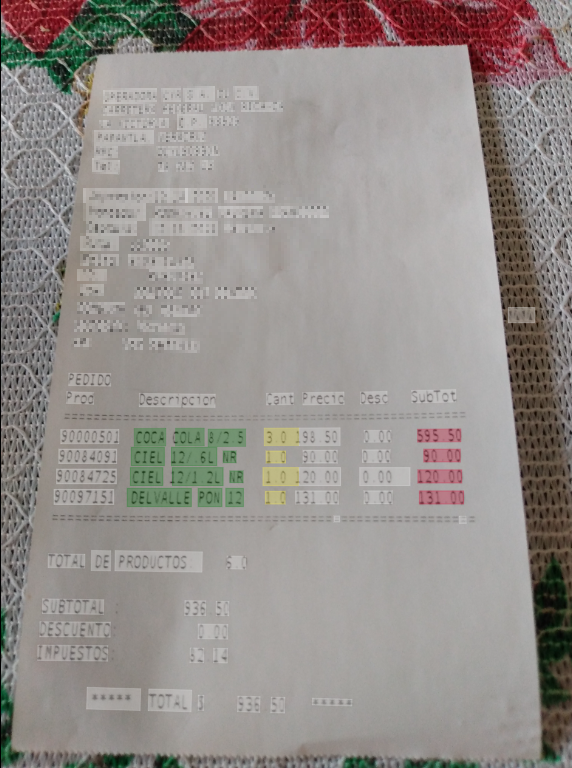}}
\subfigure[Product line grouping.]                                     
{\includegraphics[width=0.28\textwidth]{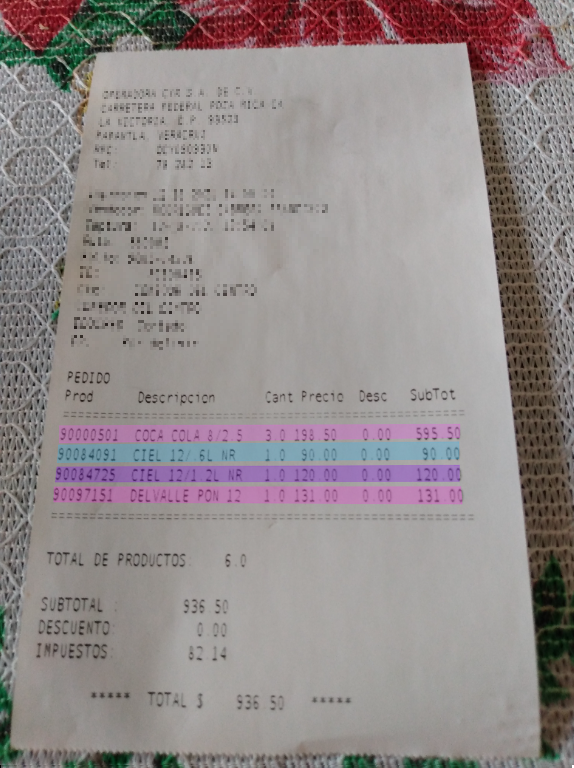}}
\subfigure[Corrected entities for codes.]                                     
{\includegraphics[width=0.28\textwidth]{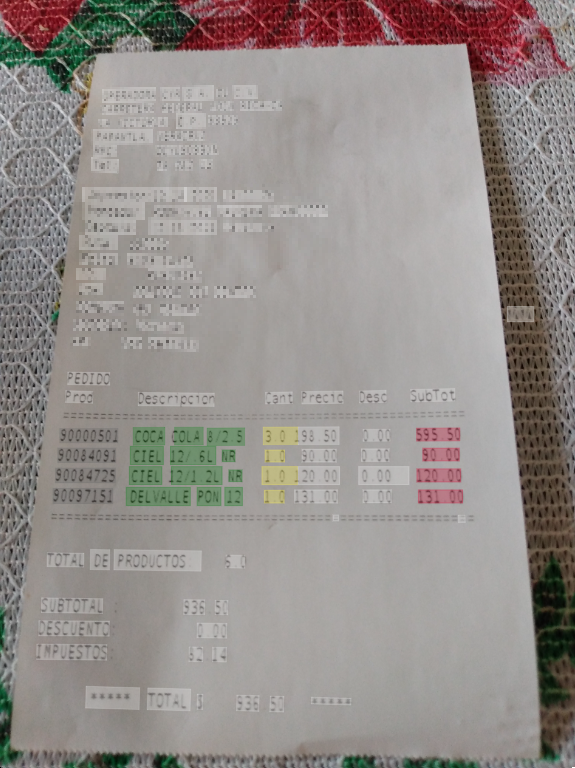}}
\caption{Visual examples of product code correction (in gray) based on rule-based heuristics.}
\label{fig:rules_example_code} 
\end{figure*}

\begin{figure*}[!ht]
\centering
\subfigure[Baseline entities.]                                     
{\includegraphics[width=0.28\textwidth]{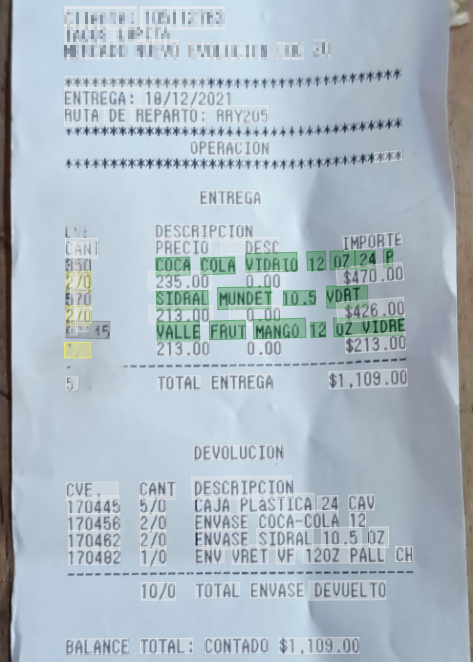}}
\subfigure[Product line grouping.]                                     
{\includegraphics[width=0.28\textwidth]{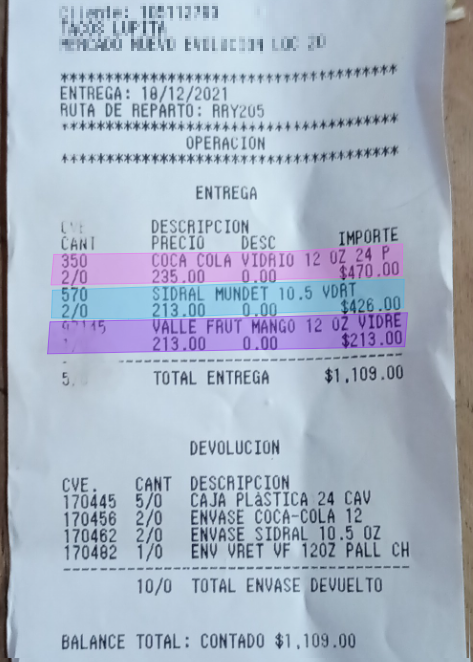}}
\subfigure[Corrected entities for prices.]                                     
{\includegraphics[width=0.28\textwidth]{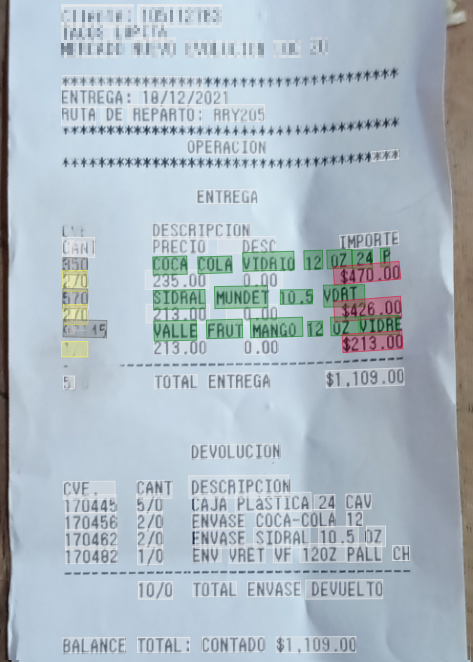}}
\caption{Visual examples of product price correction (in red) based on rule-based heuristics.}
\label{fig:rules_example_price} 
\end{figure*}

\section{Experiments}
\label{sec:experiments}

The main goal of the presented experiments is to verify the performance of our DL pipeline for KIE in purchase documents and demonstrate how \mbox{rule-based} corrections can increase final accuracy.

Initially, some public datasets are used to obtain performance metrics with the aim of comparing different entity tagging architectures with respect to our proposal based on an \mbox{encoder-decoder} built upon Transformers and CNNs, which is inspired by~\cite{ref:Yu21icpr}. The goal of this experiment is to validate that even a top \mbox{state-of-the-art} technique for entity tagging based on DL is far from perfection, so rule-based corrections can provide an added value for the most common error cases.

Besides, a dataset acquired by NielsenIQ is used to test the performance of the whole pipeline, including quantitative comparisons related to the improvements provided by rule-based corrections. Some qualitative examples from the NielsenIQ dataset can be reviewed in Figs.~\ref{fig:rules_example_code},~\ref{fig:rules_example_quantity}~and~\ref{fig:rules_example_price}.

\begin{table*}[ht!]
\centering
\begin{tabular}{|c|c|c|ccc|}
\hline
\multirow{2}{*}{\textbf{Method}} & \multirow{2}{*}{\textbf{Main reference}} & \multirow{2}{*}{\textbf{Parameters}} & \multicolumn{3}{c|}{\textbf{f1-score}}                                                    \\ \cline{4-6} 
                                 &                                     &                                      & \multicolumn{1}{c|}{\textbf{CORD}} & \multicolumn{1}{c|}{\textbf{SROIE}} & \textbf{FUNSD} \\ \hline
BERT & \citep{ref:Devlin18naacl} & 110M & \multicolumn{1}{c|}{89.7}              & \multicolumn{1}{c|}{91.0} & 60.7 \\ \hline
LayoutLMv1 & \citep{ref:Xu20kdd} & 113M & \multicolumn{1}{c|}{94.7}              & \multicolumn{1}{c|}{94.4} & 78.6 \\ \hline
LayoutLMv2 & \citep{ref:Xu21ijcnlp} & 200M & \multicolumn{1}{c|}{94.9} & \multicolumn{1}{c|}{96.2} & 82.7 \\ \hline
LayoutLMv3 & \citep{ref:Huang22arxiv} & 133M & \multicolumn{1}{c|}{96.5} & \multicolumn{1}{c|}{96.4} & 90.3 \\ \hline
LayoutXLM & \citep{ref:Xu21arxiv} & 345M & \multicolumn{1}{c|}{95.7}              & \multicolumn{1}{c|}{96.1} & 82.7 \\ \hline
SPADE & \citep{ref:Hwan21ijcnlp} & 110M & \multicolumn{1}{c|}{91.5}              & \multicolumn{1}{c|}{93.2} & 71.6  \\ \hline
BROS & \citep{ref:Hong22aaai} & 139M & \multicolumn{1}{c|}{95.3}              & \multicolumn{1}{c|}{95.5} & 81.2 \\ \hline
PICK & \citep{ref:Yu21icpr} & 68M & \multicolumn{1}{c|}{95.2} & \multicolumn{1}{c|}{96.1} & 82.5 \\ \hline
\end{tabular}
\caption{Comparison of state-of-the-art methods for entity tagging in the CORD, SROIE and FUNSD datasets.}
\label{tab:entity_tagging_results} 
\end{table*}

\subsection{Datasets}
\label{sec:datasets}

\begin{table*}[!ht]
\centering
\begin{tabular}{|c|ccccc|}
\hline
\multirow{2}{*}{\textbf{\begin{tabular}[c]{@{}c@{}}Experiments\\ for entity tagging\end{tabular}}} & \multicolumn{5}{c|}{\textbf{f1-score}}                                                                                                                                                                           \\ \cline{2-6} 
                                                                                                   & \multicolumn{1}{l|}{\textbf{Descriptions}} & \multicolumn{1}{l|}{\textbf{Codes}} & \multicolumn{1}{l|}{\textbf{Quantities}} & \multicolumn{1}{l|}{\textbf{Prices}} & \multicolumn{1}{l|}{\textbf{Whole products}} \\ \hline
With only DL                                                                                       & \multicolumn{1}{c|}{83.8}                  & \multicolumn{1}{c|}{77.8}           & \multicolumn{1}{c|}{75.5}                & \multicolumn{1}{c|}{73.1}            & 65.1                                        \\ \hline
With rule-based corrections                                                                        & \multicolumn{1}{c|}{\textbf{84.0}}         & \multicolumn{1}{c|}{\textbf{90.2}}  & \multicolumn{1}{c|}{\textbf{89.7}}       & \multicolumn{1}{c|}{\textbf{87.1}}   & \textbf{78.3}                               \\ \hline
\end{tabular}
\caption{Results in the NielsenIQ dataset for entity tagging using only DL vs adding rule-based corrections.}
\label{tab:entity_tagging_results_nielseniq} 
\end{table*}

\begin{table*}[!ht]
\centering
\begin{tabular}{|c|ccccc|}
\hline
\multirow{2}{*}{\textbf{\begin{tabular}[c]{@{}c@{}}Experiments\\ for the whole pipeline\end{tabular}}} & \multicolumn{5}{c|}{\textbf{f1-score}}                                                                                                                                                                                 \\ \cline{2-6} 
                                                                                                   & \multicolumn{1}{l|}{\textbf{Descriptions}} & \multicolumn{1}{l|}{\textbf{Codes}} & \multicolumn{1}{l|}{\textbf{Quantities}} & \multicolumn{1}{l|}{\textbf{Prices}} & \multicolumn{1}{l|}{\textbf{Whole products}} \\ \hline
With only DL                                                                                       & \multicolumn{1}{c|}{65.3}                  & \multicolumn{1}{c|}{58.9}           & \multicolumn{1}{c|}{56.9}                & \multicolumn{1}{c|}{54.6}            & 24.2                                              \\ \hline
With rule-based corrections                                                                        & \multicolumn{1}{c|}{\textbf{65.4}}         & \multicolumn{1}{c|}{\textbf{69.4}}  & \multicolumn{1}{c|}{\textbf{69.0}}       & \multicolumn{1}{c|}{\textbf{66.4}}   & \textbf{35.1}                                     \\ \hline
\end{tabular}
\caption{Results in the NielsenIQ dataset for the whole pipeline using only DL vs adding rule-based corrections.}
\label{tab:whole_pipeline_results_nielseniq} 
\end{table*}

The main characteristics of the datasets used in our experiments are briefly reviewed here. The public datasets evaluated are CORD, SROIE and FUNSD. Apart from this, we also perform specific tests for our use case related to KIE in purchase documents by means of the NielsenIQ dataset.

\subsubsection{The CORD Dataset}

The COnsolidated Receipt Dataset (CORD)\footnote{\url{http://github.com/clovaai/cord}} is focused on receipt understanding for entity tagging and linking. The dataset includes 800 receipts for the training set, 100 for the validation set and 100 for the test set. A photo and a list of OCR annotations are included for each receipt. For entity tagging, there are 30 classes related to different information from shops and restaurants in Indonesia. 

\subsubsection{The SROIE Dataset}
The Scanned Receipts OCR and Information Extraction (SROIE)\footnote{\url{http://rrc.cvc.uab.es/?ch=13}} dataset is composed of a scanned collection from 1000 store receipts. 600 images are used for training and 400 for testing. Each receipt contains around about four key text fields for entity tagging. The text annotated in the dataset mainly consists of digits and English characters.

\subsubsection{The FUNSD Dataset}

The Form Understanding in Noisy Scanned Documents (FUNSD)\footnote{\url{http://guillaumejaume.github.io/FUNSD}} dataset includes documents containing forms. It is composed of 199 scanned documents, where 9707 semantic entities are annotated from 31485 words. 149 images are used for training and 50 for testing. There are 4 semantic entities: header, question, answer and other. 

\subsubsection{The NielsenIQ Dataset}

Commonly, public datasets related to purchase document decoding include varied tags that are not always fitting the requirements of a particular use case. In our system, we want to specifically decode descriptions, codes, quantities and total prices per product. Then, we have used our own labeled data from NielsenIQ\footnote{\url{http://nielseniq.com}} considering the previously specified purchase tags to train and evaluate our full pipeline. As this dataset is composed of proprietary images, we can not publicly share them, but you can check their main characteristics in the example presented in Fig.~\ref{fig:kie_in_purchase_documents}. The images were captured with smartphones in different countries, with more than 10000 samples. Several challenges are considered due to the characteristics of data acquisition, such as different formats per retailer and country, multiple languages, or capturing quality for acquired images (e.g., varied resolutions, shadowing effects, blurring, oclussions, etc.).

\subsection{Results}
\label{sec:results}

Based on the previously described datasets, we have processed several results regarding the state of the art in entity tagging and our whole KIE pipeline. The final goal is to measure the improvements in our DL system provided by the proposed rule-based corrections.

\subsubsection{Entity Tagging Results}

In Table~\ref{tab:entity_tagging_results}, the base architectures of several approaches for entity tagging introduced in Section~\ref{sec:related_work} are compared. We analyze the number of parameters for each approach and f1-scores in the public datasets introduced for document decoding (CORD, SROIE and FUNSD).

As discussed in Section~\ref{sec:entity_tagging}, we decided to apply an entity tagging schema based on an encoder-decoder built upon Transformers and CNNs. This architecture is inspired by the PICK algorithm described in~\citep{ref:Yu21icpr}. We made this decision supported by the results presented in Table~\ref{tab:entity_tagging_results}, where PICK obtained one of the top performances in the three tested public datasets and using the lowest number of parameters (68M). This \mbox{trade-off} allows to have a high accuracy in DL predictions for entity tagging tasks jointly with light and efficient models. The average f1-score of LayoutLMv3 is the highest one and slightly better than the f1-scores obtained by PICK, but the number of parameters of LayoutLMv3 is almost double (133M). In any case, there is a margin of improvement regarding f1-scores for all the approaches that is expected to be reduced with our rule-based corrections, as will be experimented in the next section.

\subsubsection{Whole KIE Pipeline Results}

In these final experiments, we use the described NielsenIQ dataset for evaluating the improvements provided by our rule-based corrections with respect to the baseline DL predictions in KIE for purchase documents. 

In Table~\ref{tab:entity_tagging_results_nielseniq}, we present results for entity tagging including product descriptions, codes, quantities and prices. Besides, we also add a field named whole products, where a true positive represents a detected product with all its tags perfectly predicted. As can be seen, the rule-based corrections increase the performance of the mainly corrected tags, with an improvement in f1-score of about \mbox{13-14} points for codes, quantities and prices. These results clearly demonstrate how the combination of DL and rule-based approaches provides a higher performance in KIE for purchase documents.

With the aim of fully validating our proposal, we also present results for our whole pipeline in Table~\ref{tab:whole_pipeline_results_nielseniq}. This experiment includes f1-scores not only based on entity tagging, but also on OCR and product line grouping. We can see a general decrease in performance with respect to Table~\ref{tab:entity_tagging_results_nielseniq}, especially because of the more restrictive metrics calculated in this case. In particular, OCR predictions must be fully matched with the ground-truth word to consider a true positive jointly with the rest of conditions, so low quality images are greatly affected by these errors. Then, we can appreciate other challenges in performance related to KIE apart from entity tagging. In any case, the results for the whole pipeline shown in Table~\ref{tab:whole_pipeline_results_nielseniq} demonstrate again the improvements provided by rule-based corrections with respect to baseline DL predictions.

\section{Conclusions and Future Work}
\label{sec:conclusions}

The DL era has provided outstanding tools to solve complex problems in fields such as NLP and CV. Unfortunately, generalization and scalability to different use cases are still an open challenge. 

In KIE for purchase documents, DL results can be degraded in cases associated with complex formats, low resolution or multilingual scenarios, among others. However, we have demostrated how rule-based corrections can complement DL architectures to enhance the performance of document decoding in the most difficult scenarios.

We have presented a basic set of rules based on business knowledge for correcting some common purchase tags, but more general rules could be explored in the future. Besides, further works could also include confidences application to rule-based corrections or similar post-processing rules for improving other KIE tasks such as OCR  extraction.

\bibliography{anthology,custom}

\begin{thebibliography}{20}
\expandafter\ifx\csname natexlab\endcsname\relax\def\natexlab#1{#1}\fi

\bibitem[{Anwar et~al.(2022)Anwar, Khan, and Mollah}]{ref:Anwar22springer}
Nadeem Anwar, Tauseef Khan, and Ayatullah~Faruk Mollah. 2022.
\newblock {Text Detection from Scene and Born Images: How Good is Tesseract?}
\newblock \emph{Recent Trends in Communication and Intelligent Systems.
  Algorithms for Intelligent Systems, Springer}, pages 115--122.

\bibitem[{Arroyo et~al.(2022)Arroyo, Alvarez, Aller, Bergasa, and
  Ortiz}]{ref:Arroyo22mtap}
Roberto Arroyo, Sergio Alvarez, Aitor Aller, Luis~M. Bergasa, and Miguel~E.
  Ortiz. 2022.
\newblock {Fine-tuning your answers: a bag of tricks for improving VQA models}.
\newblock \emph{Multimedia Tools And Applications (MTAP), Springer}, pages
  1--25.

\bibitem[{Arroyo et~al.(2020)Arroyo, Jimenez-Cabello, and
  Martinez-Cebrian}]{ref:Arroyo20wcoling}
Roberto Arroyo, David Jimenez-Cabello, and Javier Martinez-Cebrian. 2020.
\newblock {Multi-label classification of promotions in digital leaflets using
  textual and visual information}.
\newblock In \emph{Workshop on NLP in E-Commerce (EComNLP). International
  Conference on Computational Linguistics (COLING)}, pages 11--20.

\bibitem[{Arroyo et~al.(2019)Arroyo, Tovar, Delgado, Almazan, Serrador, and
  Hurtado}]{ref:Arroyo19wcvpr}
Roberto Arroyo, Javier Tovar, Francisco~J. Delgado, Emilio~J. Almazan, Diego~G.
  Serrador, and Antonio Hurtado. 2019.
\newblock {Integration of Text-Maps in CNNs for Region Detection among
  Different Textual Categories}.
\newblock In \emph{Workshop on Language and Vision. Conference on Computer
  Vision and Pattern Recognition (CVPR)}, pages 1--4.

\bibitem[{Carbonell et~al.(2021)Carbonell, Riba, Villegas, Fornes, and
  Llados}]{ref:Carbonell21icpr}
Manuel Carbonell, Pau Riba, Mauricio Villegas, Alicia Fornes, and Josep Llados.
  2021.
\newblock {Named Entity Recognition and Relation Extraction with Graph Neural
  Networks in Semi Structured Documents}.
\newblock In \emph{International Conference on Pattern Recognition (ICPR)},
  pages 9622--9627.

\bibitem[{Devlin et~al.(2018)Devlin, Chang, Lee, and
  Toutanova}]{ref:Devlin18naacl}
Jacob Devlin, Ming-Wei Chang, Kenton Lee, and Kristina Toutanova. 2018.
\newblock {BERT: Pre-training of Deep Bidirectional Transformers for Language
  Understanding}.
\newblock In \emph{Conference of the North American Chapter of the Association
  for Computational Linguistics (NAACL-HLT)}, pages 4171--4186.

\bibitem[{He et~al.(2016)He, Zhang, Ren, and Sun}]{ref:He16cvpr}
Kaiming He, Xiangyu Zhang, Shaoqing Ren, and Jian Sun. 2016.
\newblock {Deep Residual Learning for Image Recognition}.
\newblock In \emph{Conference on Computer Vision and Pattern Recognition
  (CVPR)}, pages 770--778.

\bibitem[{Hegghammer(2022)}]{ref:Hegghammer22springer}
Thomas Hegghammer. 2022.
\newblock {OCR with Tesseract, Amazon Textract, and Google Document AI: a
  benchmarking experiment}.
\newblock \emph{Journal of Computational Social Science, Springer}, pages
  861--882.

\bibitem[{Hong et~al.(2022)Hong, Kim, Ji, Hwang, Nam, and
  Park}]{ref:Hong22aaai}
Teakgyu Hong, Donghyun Kim, Mingi Ji, Wonseok Hwang, Daehyun Nam, and Sungrae
  Park. 2022.
\newblock {BROS: A Pre-Trained Language Model Focusing on Text and Layout for
  Better Key Information Extraction from Documents}.
\newblock In \emph{AAAI Conference on Artificial Intelligence (AAAI)}, pages
  1--9.

\bibitem[{Huang et~al.(2022)Huang, Lv, Cui, Lu, and Wei}]{ref:Huang22arxiv}
Yupan Huang, Tengchao Lv, Lei Cui, Yutong Lu, and Furu Wei. 2022.
\newblock {LayoutLMv3: Pre-training for Document AI with Unified Text and Image
  Masking}.
\newblock \emph{arXiv:2204.08387}, pages 1--11.

\bibitem[{Huang et~al.(2019)Huang, Chen, He, Bai, Karatzas, Lu, and
  Jawahar}]{ref:Huang19icdar}
Zheng Huang, Kai Chen, Jianhua He, Xiang Bai, Dimosthenis Karatzas, Shjian Lu,
  and C.V. Jawahar. 2019.
\newblock {ICDAR2019 Competition on Scanned Receipt OCR and Information
  Extraction}.
\newblock In \emph{International Conference on Document Analysis and
  Recognition (ICDAR)}, pages 1526--1520.

\bibitem[{Hwang et~al.(2021)Hwang, Yim, Park, Yang, and Seo}]{ref:Hwan21ijcnlp}
Wonseok Hwang, Jinyeong Yim, Seunghyun Park, Sohee Yang, and Minjoon Seo. 2021.
\newblock {Spatial Dependency Parsing for Semi-Structured Document Information
  Extraction}.
\newblock In \emph{International Joint Conference on Natural Language
  Processing (IJCNLP)}, pages 330--343.

\bibitem[{Jaume et~al.(2019)Jaume, Ekenel, and Thiran}]{ref:Jaume19wicdar}
Guillaume Jaume, Hazim~Kemal Ekenel, and Jean-Philippe Thiran. 2019.
\newblock {FUNSD: A dataset for form understanding in noisy scanned documents}.
\newblock In \emph{Workshop on Open Services and Tools for Document Analysis.
  International Conference on Document Analysis and Recognition (ICDAR)}, pages
  1--6.

\bibitem[{Park et~al.(2019)Park, Shin, Lee, Lee, Surha, Seo, and
  Lee}]{ref:Park19wneurips}
Seunghyun Park, Seung Shin, Bado Lee, Junyeop Lee, Jaeheung Surha, Minjoon Seo,
  and Hwalsuk Lee. 2019.
\newblock {CORD: A Consolidated Receipt Dataset for Post-OCR Parsing}.
\newblock In \emph{Workshop on Document Intelligence. Conference on Neural
  Information Processing Systems (NeurIPS)}, pages 1--4.

\bibitem[{Qasim et~al.(2019)Qasim, Mahmood, and Shafait}]{ref:Qasim19icdar}
Shah~Rukh Qasim, Hassan Mahmood, and Faisal Shafait. 2019.
\newblock {Rethinking Table Recognition using Graph Neural Networks}.
\newblock In \emph{International Conference on Document Analysis and
  Recognition (ICDAR)}, pages 142--147.

\bibitem[{Vaswani et~al.(2017)Vaswani, Shazeer, Parmar, Uszkoreit, Jones,
  Gomez, Kaiser, and Polosukhin}]{ref:Vaswani17neurips}
Ashish Vaswani, Noam Shazeer, Niki Parmar, Jakob Uszkoreit, Llion Jones,
  Aidan~N. Gomez, Lukasz Kaiser, and Illia Polosukhin. 2017.
\newblock {Attention Is All You Need}.
\newblock In \emph{Conference on Neural Information Processing Systems
  (NeurIPS)}, pages 5998--6008.

\bibitem[{Xu et~al.(2021{\natexlab{a}})Xu, Xu, Lv, Cui, Wei, Wang, Lu,
  Florencio, Zhang, Che, Zhang, and Zhou}]{ref:Xu21ijcnlp}
Yang Xu, Yiheng Xu, Tengchao Lv, Lei Cui, Furu Wei, Guoxin Wang, Yijuan Lu,
  Dinei Florencio, Cha Zhang, Wanxiang Che, Min Zhang, and Lidong Zhou.
  2021{\natexlab{a}}.
\newblock {LayoutLMv2: Multi-modal Pre-training for Visually-rich Document
  Understanding}.
\newblock In \emph{International Joint Conference on Natural Language
  Processing (IJCNLP)}, pages 2579--2591.

\bibitem[{Xu et~al.(2020)Xu, Li, Cui, Huang, Wei, and Zhou}]{ref:Xu20kdd}
Yiheng Xu, Minghao Li, Lei Cui, Shaohan Huang, Furu Wei, and Ming Zhou. 2020.
\newblock {LayoutLM: Pre-training of Text and Layout for Document Image
  Understanding}.
\newblock In \emph{Int. Conference on Knowledge Discovery and Data
  Mining (KDD)}, pages 1192--1200.

\bibitem[{Xu et~al.(2021{\natexlab{b}})Xu, Lv, Cui, Wang, Lu, Florencio, Zhang,
  and Wei}]{ref:Xu21arxiv}
Yiheng Xu, Tengchao Lv, Lei Cui, Guoxin Wang, Yijuan Lu, Dinei Florencio, Cha
  Zhang, and Furu Wei. 2021{\natexlab{b}}.
\newblock {LayoutXLM: Multimodal Pre-training for Multilingual Visually-rich
  Document Understanding}.
\newblock \emph{arXiv:2104.08836}, pages 1--10.

\bibitem[{Yu et~al.(2021)Yu, Lu, Qi, Gong, and Xiao}]{ref:Yu21icpr}
Wenwen Yu, Ning Lu, Xianbiao Qi, Ping Gong, and Rong Xiao. 2021.
\newblock {PICK: Processing Key Information Extraction from Documents using
  Improved Graph Learning-Convolutional Networks}.
\newblock In \emph{International Conference on Pattern Recognition (ICPR)},
  pages 4363--4370.

\end{thebibliography}

\end{document}